\documentclass[11pt]{article}

\usepackage[]{acl}
\usepackage{times}
\usepackage{latexsym}

\usepackage[T1]{fontenc}
\usepackage[utf8]{inputenc}
\usepackage{microtype}
\usepackage{graphicx}
\usepackage{booktabs}
\usepackage{array}
\usepackage{enumitem}

\usepackage{algpseudocode}
\usepackage{tikz}
\usepackage{tikz-qtree}
\usepackage{amsfonts}

\usepackage[misc]{ifsym}
\usepackage{fontawesome5}

\usepackage[T1]{fontenc}

\usepackage[utf8]{inputenc}

\usepackage{microtype}
\usepackage{inconsolata}

\usepackage{relsize}

\usepackage{placeins}

\usepackage{stfloats}    %
\graphicspath{{figures/}}

\usepackage{xspace}

\usepackage[labelfont=bf]{caption}

\usepackage{xspace}

\newcommand{\studio}{\texttt{string2string Studio}}
\newcommand{\lib}{\texttt{string2string}\xspace}

\title{\studio{}: \\An Interactive, In-Browser Platform for String-to-String Algorithms}

\author{
Mirac Suzgun\\
Stanford University \\
\And
James Zou\\
Stanford University \\
\And
Stuart M. Shieber\\
Harvard University \\
\And
Dan Jurafsky\\
Stanford University \\
}

\begin{document}
\maketitle

\begin{abstract}
We present \textbf{\studio}, an interactive in-browser platform for string-to-string analysis across natural language processing, computational biology, and the digital humanities. The system integrates six main modules (alignment, distance, similarity, search, generation metrics, and \textsc{blast} homology search), operating at character, word, token, line, and residue levels. Its C++-based algorithms compile to WebAssembly, so core operations run locally by default without any installation or data upload. The interface reports scores with their ``evidence'' (alignments, edit paths, metric matches, search hits, and homology traces), making methods inspectable, debuggable, and comparable on shared inputs. Internal benchmarks show speedups of up to $2{,}500\times$ over the Python predecessor, faster global/local alignment than a general-purpose native C aligner, and exact agreement with independent references under declared settings. For homology search, the scoped client-side \texttt{blastn} path closely matches \textsc{ncbi} \textsc{blast}+ rankings and statistics under matched parameters. A curated showcase and \emph{Learn} mode present canonical algorithms and metrics as reusable demonstrations. \studio{} is open-source and freely available at \url{string2string.org}.\footnote{\ Correspondence to: \url{msuzgun@stanford.edu}.}
\end{abstract}

\begin{figure*}[t]
  \centering
  \includegraphics[width=\textwidth]{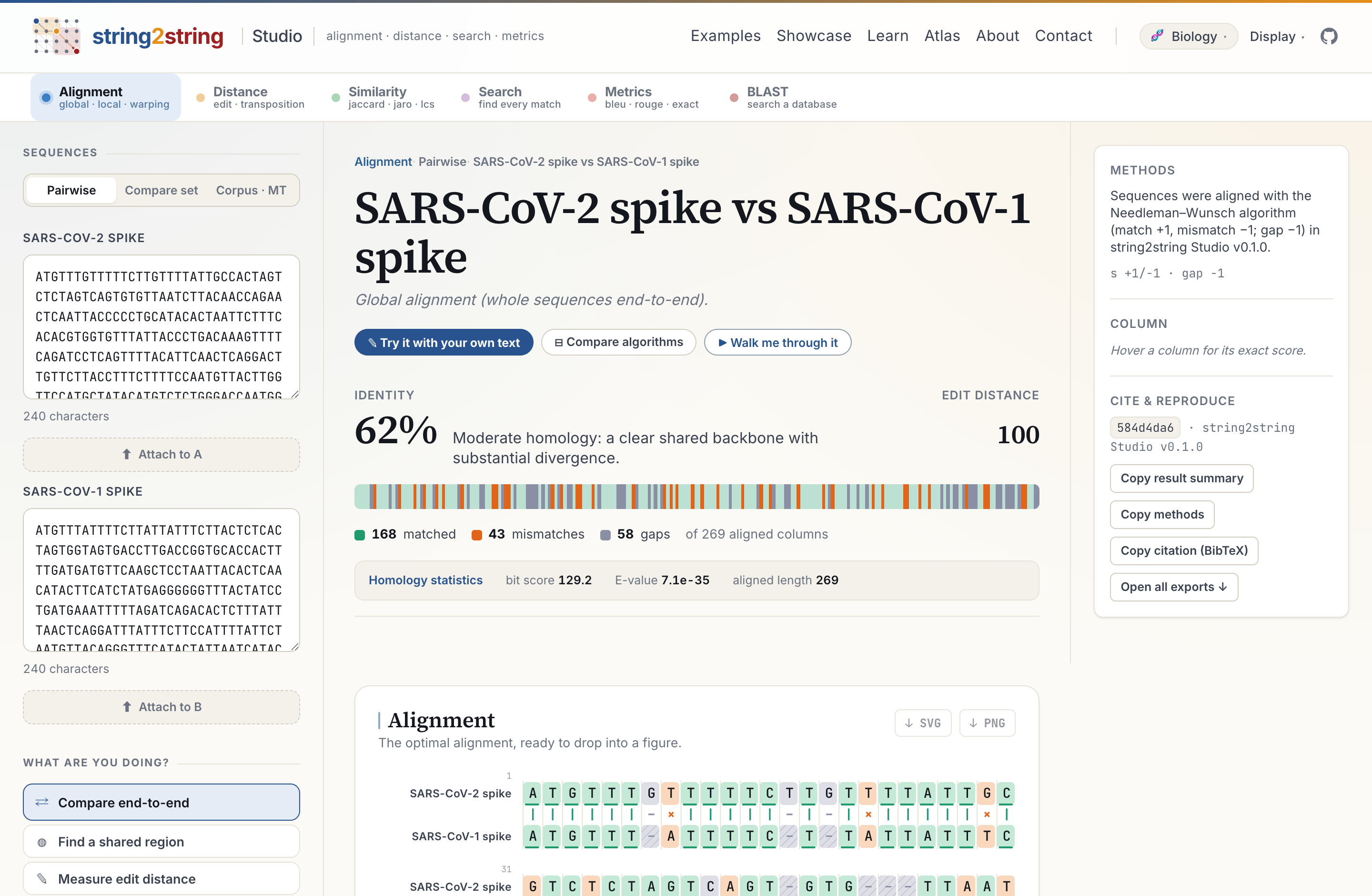}
  \caption{The \studio{} workbench. The algorithm family is selectable across the top (alignment, distance, similarity, search, metrics, \textsc{blast}). Here, we illustrate the global alignment of a pair of coronavirus spike genes: the header reports identity and edit distance, and the identity track marks each of the 269 aligned columns as a match, substitution, or gap. The right panel records the method, citation, and export options; the interface adapts defaults and terminology to the user's field. Core computations run locally in the browser by default.}
  \label{fig:teaser}
\end{figure*}

\section{Introduction}
String-to-string algorithms appear across many areas of computing. In natural language processing, they underlie spelling correction, near-duplicate and training-data contamination detection, and the evaluation of machine-translated and other generated texts. In computational biology, they align sequences, search for chromoses , and reconstruct phylogenies. In the digital humanities, they collate manuscript contents and track how texts change across editions and translations. While these applications differ in data type, scale, and convention, they all rely on a common set of operations: aligning strings, measuring differences, computing similarity, searching for exact/approximate matches, and scoring sequence-level outputs \citep[][\emph{inter alia}]{levenshtein1966binary,needleman1970general,gusfield1997algorithms,jurafsky2025slp}.

Despite this shared foundation, the tools for these operations, however, remain mostly scattered. Some tools are distributed as programming libraries; others as command-line tools, web services, or domain-specific applications. Many are pretty powerful, but they often make it difficult to compare methods on the same inputs, inspect intermediate structures, adjust parameters interactively, or move between textual and biological use cases. Our previous \lib{} library \citep{suzgun2024string2string} sought to address part of this fragmentation by collecting many string-to-string algorithms under a unified Python API. But a typical library is not an interactive environment: it does not, on its own, let users see an alignment, inspect an edit path, compare metrics, export a visualization, or run analyses locally in the browser.

Here, we present \studio{}, an interactive browser-based platform for string-to-string analysis (Figure~\ref{fig:teaser}). The system is organized around six modules: (a) \emph{alignment}, (b) \emph{distance}, (c) \emph{similarity}, (d) \emph{search}, (e) \emph{metrics}, and (f) \textsc{blast} \emph{homology search}. It supports multiple granularities, including character, word, token, line, and residue level. Its computational engine is implemented in C++ and compiled to WebAssembly \citep{haas2017bringing}; thus, the main operations run on the client, with no software to install and no data leaving the machine by default. The interface presents outputs as linked, inspectable, and exportable objects, enabling users to compare methods on the same inputs and examine the structure behind the scores.

This work makes four contributions. \textit{First}, we introduce, to our knowledge, the first browser-based workbench that integrates these methods in a single interactive environment without installation or programming, and that makes their outputs entirely inspectable---exposing alignments, edit paths, metric matches, search hits, and homology-search traces as linked visual objects rather than opaque scalars. \textit{Second}, we describe a C++/WebAssembly engine for interactive client-side use that runs 100--2{,}500 times faster than its Python predecessor, outpaces a general-purpose native C aligner on global and local alignment, and agrees exactly with independent reference implementations. \textit{Third}, we add a scoped client-side \texttt{blastn} path for homology search and evaluate its statistics against \textsc{ncbi} \textsc{blast}+ under matched parameters. \textit{Fourth}, we provide a curated showcase and illustrations that packages classical and contemporary methods as executable, shareable illustrations for teaching, debugging, and cross-domain comparison (Figure~\ref{fig:gallery}).

\section{Related Work}
\label{sec:related}

Our platform draws on several families of established tools \& platforms. General-purpose libraries, including our predecessor \citep{suzgun2024string2string}, \href{https://github.com/life4/textdistance}{\texttt{textdistance}}, \href{https://github.com/jamesturk/jellyfish}{\texttt{jellyfish}}, and \href{https://github.com/rapidfuzz/RapidFuzz}{\texttt{RapidFuzz}}, and optimized kernels, including \href{https://github.com/Martinsos/edlib}{\texttt{edlib}} \citep{sosic2017edlib}, \href{https://github.com/mengyao/complete-striped-smith-waterman-library}{\textsc{ssw}} \citep{zhao2013ssw}, \href{https://github.com/jeffdaily/parasail}{\texttt{parasail}} \citep{daily2016parasail}, and \href{https://biopython.org/}{\texttt{Biopython}}'s aligner \citep{cock2009biopython}, provide implementations through programming interfaces. Domain-tailored systems naturally serve more specialized needs: \href{https://blast.ncbi.nlm.nih.gov/Blast.cgi}{\textsc{NCBI BLAST}} \citep{altschul1990basic,camacho2009blast} and \href{https://www.ebi.ac.uk/jdispatcher/emboss}{\textsc{emboss}} for sequence search, \href{https://www.jalview.org/}{Jalview} \citep{waterhouse2009jalview} for alignment display and editing, \href{https://collatex.net/}{CollateX} and Juxta for manuscript collation, and sacre\textsc{bleu} \citep{post2018call}, jiwer, and \href{https://huggingface.co/docs/evaluate/v0.4.5/en/index}{Hugging Face \texttt{evaluate}} for generation metrics. Interactive analysis systems in NLP, including VizSeq \citep{wang2019vizseq}, compare-mt \citep{neubig2019comparemt}, ExplainaBoard \citep{liu2021explainaboard}, MT-ComparEval \citep{klejch2015mtcompareval}, MT-Telescope \citep{rei2021mttelescope}, the Language Interpretability Tool \citep{tenney2020lit}, BERTViz \citep{vig2019bertviz}, exBERT \citep{hoover2020exbert}, Seq2Seq-Vis \citep{strobelt2019seq2seqvis}, and the Annotated Transformer \citep{rush2018annotated}, show the value of making model and metric behavior visible. Browser-based scientific computation has recent precedent in bioinformatics \citep{biowasm,priyam2019sequenceserver,ji2024viralwasm}.

\begin{table*}[!ht]
\centering
\small
\setlength{\tabcolsep}{5.5pt}
\resizebox{\textwidth}{!}{%
\begin{tabular}{@{}lccccccc@{}}
\toprule
System & Breadth & Multi-level & Interactive & Cross-method & Verified & In-browser & Speed \\
\midrule
\lib{} library \citep{suzgun2024string2string}           & $\bullet$ & $\bullet$ & $\circ$   &           &           &           & $\circ$   \\
\texttt{edlib} \citep{sosic2017edlib}                    &           &           &           &           &           &           & $\bullet$ \\
\texttt{RapidFuzz}                                       &           & $\circ$   &           &           &           &           & $\bullet$ \\
Biopython aligner \citep{cock2009biopython}              & $\circ$   &           &           &           &           &           & $\circ$   \\
\textsc{ncbi} \textsc{blast} \citep{camacho2009blast}    & $\circ$   &           & $\circ$   &           & $\bullet$ & $\circ$   & $\bullet$ \\
sacre\textsc{bleu} \citep{post2018call}                  &           & $\circ$   &           &           & $\bullet$ &           & $\bullet$ \\
Jalview \citep{waterhouse2009jalview}                    & $\circ$   &           & $\bullet$ &           &           & $\circ$   &           \\
compare-mt\,/\,VizSeq \citep{neubig2019comparemt,wang2019vizseq} &   & $\circ$   & $\bullet$ & $\circ$   &           & $\circ$   &           \\
\midrule
\textbf{\studio{} (this work)}                           & $\bullet$ & $\bullet$ & $\bullet$ & $\bullet$ & $\bullet$ & $\bullet$ & $\circ$   \\
\bottomrule
\end{tabular}%
}
\caption{Representative tools by property ($\bullet$~present, $\circ$~partial, blank~absent).
\textbf{Breadth}: spans several classes of the family (alignment, distance, similarity, search).
\textbf{Multi-level}: runs at the character, word, token, and line level ($\bullet$) or a subset
($\circ$). \textbf{Interactive}: results as interactive figures ($\bullet$) or static plots/reports
($\circ$). \textbf{Cross-method}: several methods on one input in one interface. \textbf{Verified}: is,
or is validated bit-for-bit against, a recognized reference. \textbf{In-browser}: computes client-side
with no install ($\bullet$), or a browser build is a transpiled port or server-backed ($\circ$).
\textbf{Speed}: competitive with native code. \texttt{edlib} and \texttt{RapidFuzz} are faster on single
primitives and \textsc{ncbi} \textsc{blast} searches at database scale; \studio{} is alone in combining
breadth and multi-level operation with a verified, cross-method, client-side engine.}
\label{tab:compare}
\end{table*}

The gap is \emph{not} the absence of algorithms but the absence of a comprehensive environment around them: the fastest tools are usually narrow primitives, broad tools are libraries, and interactive tools are usually tied to one domain or backed by a server. Browser genomics projects show that native bioinformatics software can run on the web, but they expose genomics command-line tools rather than a cross-domain workbench, and systems like SequenceServer \citep{priyam2019sequenceserver} keep the computation server-side. \studio{} thus targets a different point in its design space: a verified, client-side environment in which these heterogeneous algorithms are executable, inspectable, and shareable as reusable illustrations rather than isolated library calls or opaque scores (Table~\ref{tab:compare}).

\section{The Platform: \studio{}}
\label{sec:platform}

We have organized our Studio around six core modules. The \textbf{\emph{alignment}} module includes global, local, semi-global, affine-gap, linear-space Hirschberg \citep{hirschberg1975linear}, banded, and dynamic-time-warping alignments. The \textbf{\emph{distance}} module includes Levenshtein, Damerau-Levenshtein, Hamming, Jaro-Winkler, and longest-common-subsequence distances. The \textbf{\emph{similarity}} and \textbf{\emph{search}} modules cover set- and vector-based similarity, exact and approximate lexical search, $k$-mismatch search, \textsc{iupac}-degenerate search, and both-strand sequence search. The \textbf{\emph{metrics}} module implements generation-evaluation measures, including \textsc{bleu} \citep{papineni2002bleu}, chr\textsc{f}/chr\textsc{f}++ \citep{popovic2015chrf}, and \textsc{rouge}-1/2/\textsc{l} \citep{lin2004rouge}, with single- and multi-reference inputs and paired-bootstrap significance. The \textbf{\emph{homology}} module implements the client-side \textsc{blast} workflow described in \S~\ref{sec:blast}. Following \lib{}~\citep{suzgun2024string2string}, these methods run at multiple levels (e.g., character, word, token, line, and residue), so the same interface can compare token sequences, prose passages, biological sequences, and textual witnesses.\footnote{\ Table~\ref{tab:taxonomy} lists the implemented methods, levels, and views.}

The modules share one result schema. A computation returns the input level, parameters, scalar summaries, and the evidence behind those summaries: an alignment and path for alignment algorithms, an edit script for distances, matched spans for search, sufficient statistics for corpus metrics, or seeds, extensions, bit-scores, and $E$-values for homology search. This schema was a core design choice behind the interface. Because every result has the same basic shape (that is, score, evidence, parameters, and exports), the platform can easily compare several methods on one input, attach each number to the structure that produced it, reproduce a result from a link, and validate outputs against independent implementations.

The platform runs \emph{locally} by default: algorithms execute in the browser, so users install nothing and their text or sequences stay on the machine unless they choose a remote database search. Local execution also makes interaction practical: changing a tokenization level, gap penalty, scoring scheme, or metric option recomputes the result without a server round-trip. The interface adapts defaults, terminology, and examples to the user's field, but the same methods remain available across fields.

Our showcase is dynamic: each curated example stores the inputs, level, parameters, result evidence, citation, and export profile needed to reproduce a computation, and starts from a concrete task---near-duplicate and plagiarism detection, translation and textual-witness comparison, text-generation evaluation, or genome comparison (e.g., the \textsc{sars}-\textsc{cov}-1 and \textsc{sars}-\textsc{cov}-2 spike genes)---so that users can change assumptions or parameters and watch the evidence change. A ``publication'' feature renders results as self-contained figures with citations and exports, so the same object can be inspected in the browser, shared by link, or moved into an article.

\begin{figure*}[t]
  \centering
  \includegraphics[width=\textwidth]{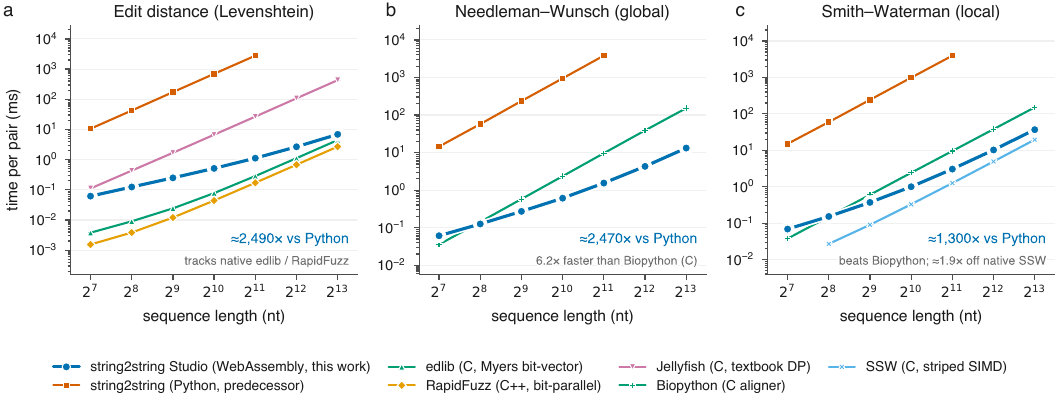}
  \caption{Runtime vs. sequence length (log--log; medians over $30$ random pairs per length, Apple M1 Pro). Each panel covers one algorithm: edit distance (left), global alignment (center), and local alignment (right). Our Python predecessor~\citep{suzgun2024string2string} was measurable only to length $2^{11}$; the annotated factors are read at length $2048$. For global and local alignment, our C++/WebAssembly engine is faster than Biopython's general-purpose C aligner and within about a factor of two of the \textsc{simd}-C \textsc{ssw} library, with identical scores.}
  \label{fig:perf}
\end{figure*}

\section{Engine}
\label{sec:engine}
While designing our Studio, we sought to satisfy four factors at once: \emph{precision}, \emph{breadth}, \emph{efficiency}, and \emph{locality}. For instance, a server wrapper around our Python library would simplify engineering but weaken privacy and make interaction depend on network latency; similarly, a pure TypeScript implementation would be easier to inspect but too slow for long alignments and repeated recomputation. 
We separate the system into a typed algorithmic core, a browser dispatcher that selects the fastest available backend, and view components that consume the same result objects across domains.

\textbf{Implementation.}
The core algorithms are reimplemented in C++ and compiled to two WebAssembly builds: a 128-bit \textsc{simd} build and a scalar fallback, selected at load time by feature detection. A separate TypeScript implementation follows the same specifications and serves both as documentation and as an independent reference. The engine aligns inputs too large for the full quadratic-space matrix with linear-space Hirschberg in a background worker, preserving optimality while keeping the interface responsive. The engine returns typed results rather than display-specific strings, so the same output can be rendered as an alignment, edit path, table, chart, export file, or record.

\textbf{Speed.}
We benchmark our engine against the Python \lib{} library and specialized native libraries; all measurements are medians over $30$ random sequence pairs per length on an Apple M1 Pro (\textsc{arm64}).
Against the Python predecessor, our engine is $100$ to $2500$ times faster, 
with the largest measured speedups at length $2048$: roughly $2500\times$ for edit distance and global alignment, and $1300\times$ for local alignment. \emph{See} Figure~\ref{fig:perf}.
These gains reflect both implementation and algorithmic changes, including bit-parallel \citep{myers1999fast} and \textsc{simd} kernels. Against native C baselines, our striped Smith--Waterman returns identical scores to \textsc{ssw} \citep{zhao2013ssw} and runs within a factor of $1.8$ at length $16{,}384$, while our anti-diagonal \textsc{simd} Needleman-Wunsch is $6.2\times$ faster than Biopython's general-purpose C aligner at length $2048$.\footnote{\ These comparisons are deliberately scoped. Specialized bit-parallel libraries such as \texttt{edlib} and \texttt{RapidFuzz} remain faster for edit distance, and we did not obtain a native \textsc{simd}-C global aligner on \textsc{arm64} because \texttt{parasail} did not compile. Our claim is therefore not that \studio{} is the fastest implementation of every primitive, but that a broad, zero-install browser engine can be fast enough for interactive analysis while remaining close to optimized native code on representative alignment tasks.}

\textbf{Space \& verification.} The full DP (dynamic programming) matrix, needed only when the user asks to inspect the whole table, exceeds the 2\textsc{gb} WebAssembly heap at length 12K. Linear-space methods return the same optimum to length 40K, though arbitrarily very long inputs may take seconds rather than feel instantaneous. Verification follows the schema described above: each algorithm is checked against the \lib{} library and the TypeScript reference at char- and token-levels; \textsc{simd} kernels are checked against scalar kernels over parameter grids; generation metrics are checked against their named reference implementations; and the \texttt{blastn} path is checked against \textsc{ncbi} \textsc{blast}+ under matched settings (\S\ref{sec:eval}).

\begin{figure*}[t]
  \centering
  \includegraphics[width=\textwidth]{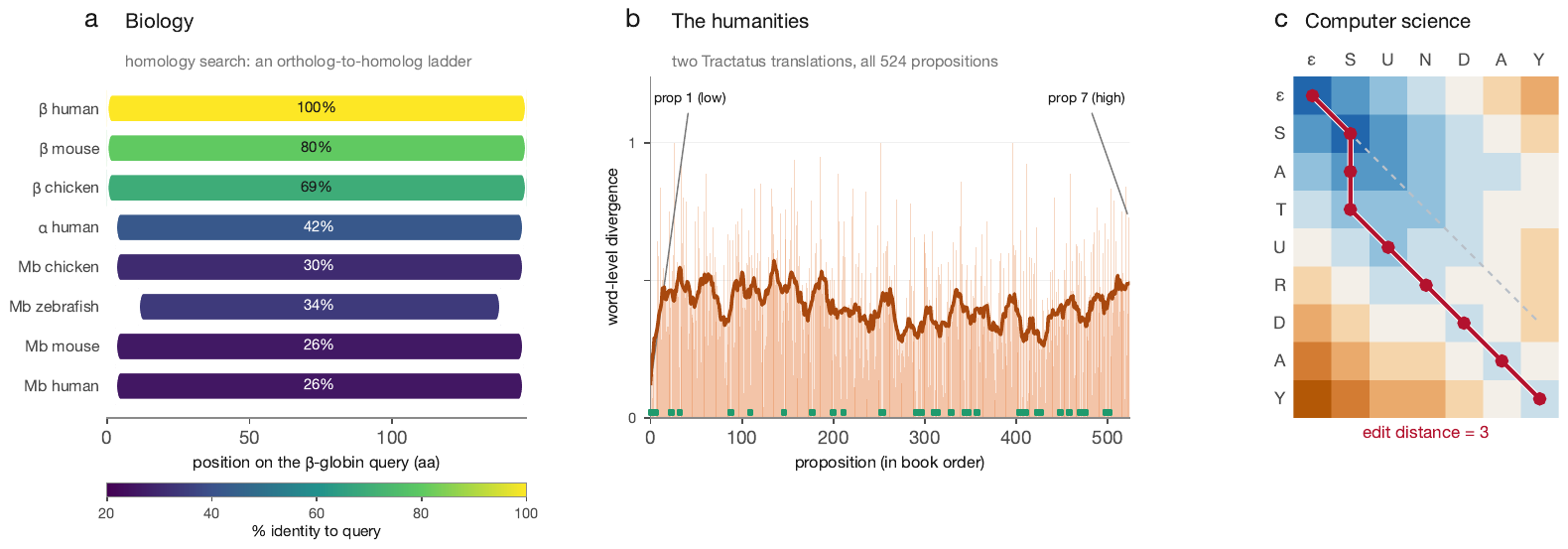}
  \caption{The same verified engine, rendered in the idiom of three disciplines. (a)~{Biology}: a human $\beta$-globin query against a bundled landmark database returns a graded homology ladder---orthologs (mouse, chicken), the $\alpha$-globin paralog, and distant myoglobin homologs---each bar colored by \% identity. (b)~{Humanities}: word-level divergence between two English \emph{Tractatus} translations across all 524 propositions (faint bars) with a rolling-mean trend (line); the $35$ identical propositions ($7\%$) are marked green. (c)~{Computer science}: the edit-distance cost field with the optimal path traced through it, each off-diagonal step an insertion or deletion.}
  \label{fig:gallery}
\end{figure*}

\section{Visualization, Teaching, and Interaction}
\label{sec:viz}

The Studio treats visualization as main part of the offerings, not as a second-thought. Each main result is rendered as a set of linked views (Figure~\ref{fig:gallery}). Selecting a column or span in one view highlights the corresponding evidence in the others and in a docked inspector; changing a gap penalty or tokenization level re-flows the result in place; zooming moves from an overview to individual symbols; and comparison views place several methods on the same input (for example, Levenshtein, Damerau--Levenshtein, and global alignment side by side). These interactions depend on the engine's local recomputation and on the shared result schema.

The same computation can be rendered in different disciplinary idioms. An alignment can appear as a per-column identity track (Figure~\ref{fig:teaser}), a dot plot, a synteny ribbon, an edit script, a version diff, a six-frame translation for coding sequence, a critical apparatus, or the full dynamic-programming matrix with the optimal path. For text-generation evaluation, \studio{} computes corpus-level \textsc{bleu} and chr\textsc{f} from summed sufficient statistics, allowing corpus-score differences to be attributed to individual segments without re-scoring subsets; a segment can then be opened to the matches and edits the metric rewards or penalizes, with a paired-bootstrap confidence interval on the system-level difference \citep{koehn2004statistical}. The point is not to replace expert judgment with a score, but to make the score more auditable.

Every view exports to the formats its audience might expect: domain interchange formats such as \textsc{fasta}, \textsc{clustal}, \textsc{cigar}, \textsc{paf}, \textsc{vcf}, and a critical apparatus, as well as \textsc{pdf}, \textsc{csv}, \textsc{json}, \textsc{svg}, and \LaTeX{}. Each figure also carries a reproducible link containing the relevant inputs and parameters. The same mechanism supports the public showcase and the ``\emph{Learn}'' mode: users can step through the dynamic program behind edit distance or alignment, alter the cost model, and see how the matrix, path, alignment, and final score change together.

\section{Homology Search in the Browser}
\label{sec:blast}

The original \lib{} paper originally identified \textsc{blast} support as future work \citep{suzgun2024string2string}. We are very delighted to report that \studio{} finally delivers it, adding a scoped homology-search path while preserving the platform's default locality. For a query sequence, the engine follows the familiar seed-and-extend structure of \textsc{blast}: it indexes short seeds, finds candidate seed hits, extends them into local alignments, and reports bit-scores and $E$-values computed from Karlin-Altschul statistics \citep{karlin1990methods,altschul1990basic,altschul1997gapped,camacho2009blast}. Because the output uses the same result schema as the rest of the platform, the user can inspect not only the final ranked hits but also the seeds, extensions, scores, parameters, and aligned regions behind them.

The module currently supports three modes. Searches can run against bundled/customized landmark databases, against a user-supplied database loaded locally, or, when the user explicitly chooses it, against remote \textsc{ncbi}. Local search lets a user screen an unpublished or potentially sensitive sequence without transmitting it; the optional remote path is appropriate when the relevant database is too large to ship with the browser. The result view mirrors familiar biological reports (a ranked hit table, per-query hit distribution, conservation skyline, and aligned hit ladder) while adding linked inspection and export. Figure~\ref{fig:homology} shows the browser report for a human $\beta$-globin query.

We also note that \studio{} does not replace \textsc{ncbi} \textsc{blast}+ at database scale, and we do not claim full parity across all \textsc{blast} tasks, databases, or scoring regimes; the module's role here is to bring local, inspectable homology search into the same cross-domain workbench as the other methods, with its \texttt{blastn} statistics evaluated in \S~\ref{sec:eval}.

\section{Evaluation}
\label{sec:eval}
In every stage of our implementation, we have carefully tested our algorithms in terms of time \& space against other established frameworks and libraries. Here, we briefly discuss our evaluation criteria and results. We shall remark that our engine is fast, efficient, and reliable for real-time interactions; that our results match declared references; and that our client-side \texttt{blastn} statistics overall agree with \textsc{ncbi} \textsc{blast}+ under our setting. 

\textbf{Runtime and memory.}
\S~\ref{sec:engine} and Figure~\ref{fig:perf} compare runtime against the Python \lib{} library and against \texttt{edlib}, \texttt{RapidFuzz}, Biopython, and native \textsc{ssw} across lengths $2^7$ to $2^{13}$. All systems use the same inputs, hardware, and repetition protocol, and baselines are called through their recommended interfaces. We also report the input length at which the full dynamic-programming matrix exhausts the WebAssembly heap, because displaying the matrix is a stricter memory requirement than returning only the optimal score or alignment.

\textbf{Reference agreement.}
Every core algorithm is compared against two independent implementations, namely the \lib{} library and the TypeScript reference, at both character and token levels. \emph{See also} Appendix~\ref{app:tables}. The generation metrics are checked against the reference named in the interface: \textsc{bleu} against \textsc{nltk} method~1, chr\textsc{f}/chr\textsc{f}++ against sacre\textsc{bleu}, and \textsc{rouge}-1/2/\textsc{l} against \texttt{rouge-score}. Under the stated settings, the differences are exactly $0$. We therefore report agreement with declared implementations and configurations; the tokenization and smoothing signature are recorded for reproducibility \citep{post2018call}.

\textbf{\textsc{blast} concordance.}
For homology search, we ran the client-side \texttt{blastn} implementation against \textsc{ncbi} \textsc{blast}+~2.17 on a six-sequence 16S r\textsc{rna} database, using identical query sequences and classic-\texttt{blastn} parameters. The hit ranking is identical across the two systems; bit-scores differ by at most $0.43\%$; percent identity differs by at most $0.18$ points; and primary-\textsc{hsp} $E$-values agree to the same order of magnitude, with differences expected from implementation choices and numerical flooring for extremely small values. Appendix~\ref{app:tables} gives the database, queries, hits, and parameters.

\section{Limitations}

We highlight five main limitations. First, browser memory constrains the largest inspectable dynamic programs: the full-matrix view is unavailable beyond length $12{,}000$, though linear-space methods return the same optimum on longer inputs. Second, the metric suite is deliberately lexical and symbolic. Neural and semantic metrics such as \textsc{bertscore} are outside the present local, zero-installation client. We hope to deploy neural-based approaches in the future. Third, the homology-search evaluation is intentionally narrow: bundled databases are illustrative, large-scale database search is delegated to an optional remote call, and only \texttt{blastn} statistics are validated against \textsc{ncbi} \textsc{blast}+ here. Fourth, the showcase and \emph{Learn} mode are useful and insightful design contributions, but we have not evaluated learning interventions and efficacy: we have not run a formal user study and make no measured claim about learning gains, usability, or analyst productivity. Fifth, the examples in our Studio are weighted toward English and Latin-script text for now, and the runtime benchmark uses random sequence pairs; broader multilingual and similarity-stratified evaluations remain future work.

\clearpage

\section*{Acknowledgments}
Our initial string2string library owes a debt of gratitude to the following individuals for their contributions, comments, and feedback: Federico Bianchi, Corinna Coupette, Sebastian Gehrmann, Tayfun Gür, Şule Kahraman, Deniz Keleş, Luke Melas-Kyriazi, Christopher Manning, Tolúlopé Ògúnrèmí, Alexander ``Sasha'' Rush, Kyle Swanson, and Garrett Tanzer.  We shall also thank Deniz Keleş, Kyle Swanson, and Filippos Sytilidis for their helpful comments and suggestions. Suzgun also wishes to thank Sebastian Gehrmann, whose conversations and insights on HCI design principles many years ago helped shape the design and organization of the website. Suzgun gratefully acknowledges the support of a Google PhD Fellowship.

\section*{Ethics Statement}

By default, all the computations performed are local: analyzed text and sequences remain on the user's machine unless the user explicitly invokes a remote database search. This design is important for unpublished biological sequences, sensitive corpora, and classroom or research settings where users may not have permission to upload data.  

\textsc{blast} is a trademark of the National Library of Medicine (NLM). Our implementation is independent and not endorsed by \textsc{ncbi}, the optional remote path is opt-in and attributed, and we cite the algorithm and statistics it reproduces \citep{altschul1990basic,karlin1990methods,camacho2009blast}. Any usage analytics are limited and disclosed. The Studio is released under a permissive open-source license.

\section*{Disclaimer} 

Portions of this manuscript were edited with the assistance of AI chatbots, used primarily to condense selected passages and improve the overall clarity, style, and flow. Every model-generated suggestion was reviewed by the authors, and only revisions limited to the word and sentence level were ultimately adopted; all substantive content, claims, and analyses remain the authors’ sole responsibility.

\bibliography{custom}

\clearpage
\appendix

\onecolumn
\section{Supplementary Figures and Tables}
\label{app:tables}

\begin{figure}[!htbp]
  \centering
  \includegraphics[width=0.90\textwidth]{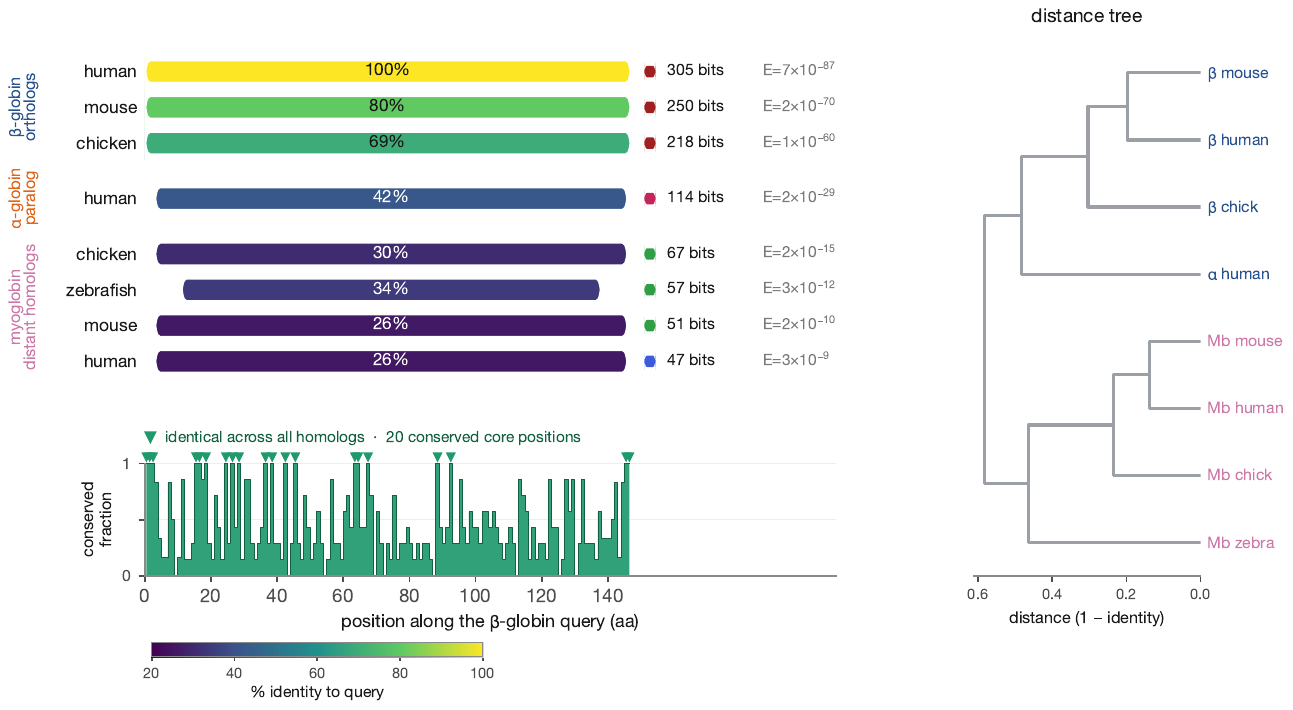}
  \caption{\emph{Homology search in depth.} One human $\beta$-globin query, searched entirely in the browser against a $29$-sequence landmark database, recovers the globin superfamily as a graded ladder: $\beta$-globin orthologs (mouse $80\%$, chicken $69\%$ identity), the $\alpha$-globin paralog ($42\%$), and distant myoglobin homologs ($26$--$34\%$), each bar coloured by percent identity and annotated with its bit-score and $E$-value. Below, the per-residue conservation skyline marks the $20$ positions identical across all homologs; at right, the distance tree built from the same alignments recovers the globin and myoglobin clades. The report exposes the Karlin--Altschul parameters ($\lambda$, $K$) and $E$-values; quantitative concordance with \textsc{ncbi} \textsc{blast}+ is evaluated for \texttt{blastn} in Figure~\ref{fig:verify}.}
  \label{fig:homology}
\end{figure}

\begin{figure}[!htbp]
  \centering
  \includegraphics[width=0.90\textwidth]{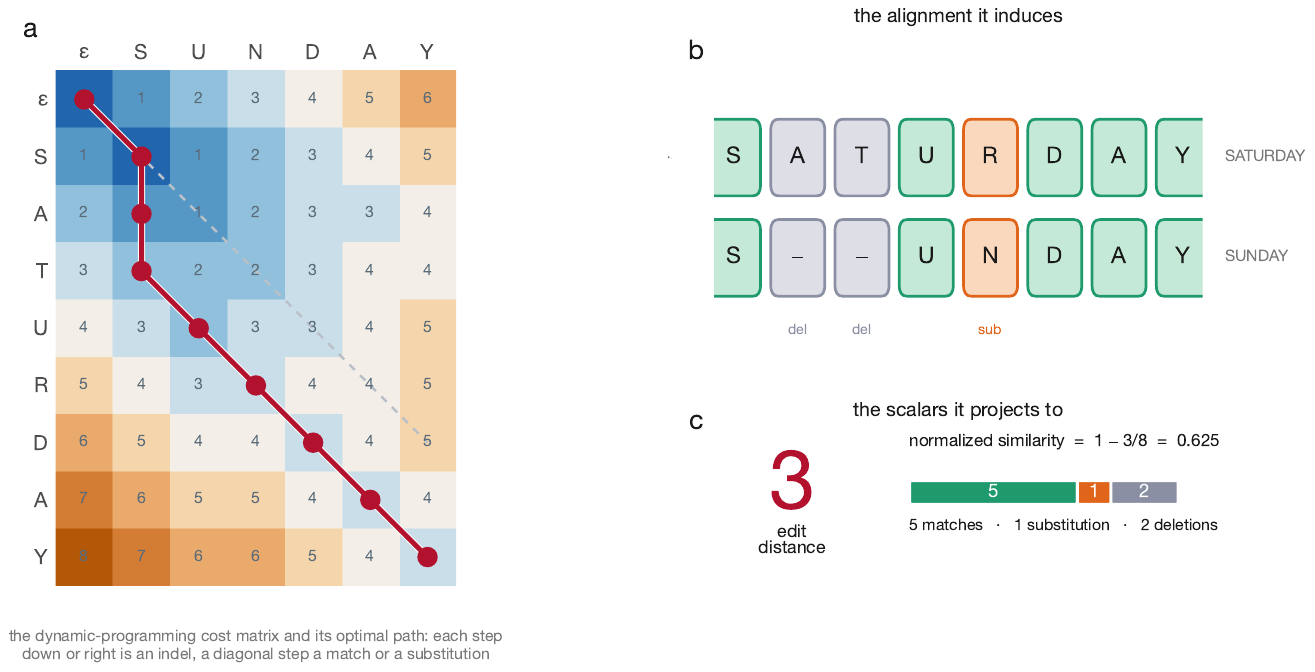}
  \caption{\emph{One alignment, three lenses.} A reported edit distance is a projection of a larger object. For \texttt{SATURDAY} vs \texttt{SUNDAY}, the same computation is shown as (a)~the dynamic-programming cost matrix with the optimal path traced through it, (b)~the alignment that path induces, column by column (five matches, one substitution, two deletions), and (c)~the scalars it projects to, including the edit distance of $3$. Every metric in the suite is one such projection.}
  \label{fig:lenses}
\end{figure}

\newpage

\begin{figure}[h]
  \centering
  \includegraphics[width=\columnwidth]{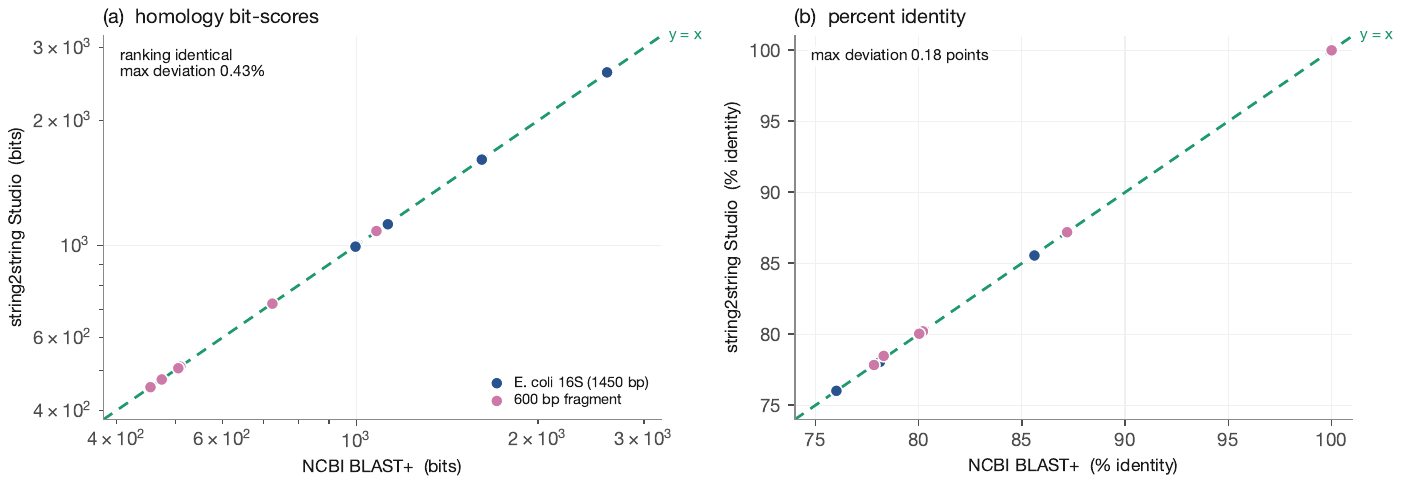}
  \caption{\emph{Client-side \texttt{blastn} is faithful under the evaluated setting.} Our
  \texttt{blastn} run is compared with \textsc{ncbi} \textsc{blast}+ 2.17 on a real 16S r\textsc{rna} database
  with identical parameters: (a)~bit-scores and (b)~percent identity for every hit lie on the line of equality
  (ranking is identical; bit-scores agree within $0.43\%$ and identity within $0.18$ points). The rest of
  the engine's exact agreement with its reference implementations (\textsc{nltk}, sacre\textsc{bleu},
  \texttt{rouge-score}, and bit-exact parity with the \lib{} library) is reported in \S~\ref{sec:eval}.}
  \label{fig:verify}
\end{figure}

\begin{table*}[!h]
\centering
\small
\setlength{\tabcolsep}{8pt}
\renewcommand{\arraystretch}{1.18}
\resizebox{\textwidth}{!}{%
\begin{tabular}{@{}lll@{}}
\toprule
 & \lib{} library \citep{suzgun2024string2string} & \textbf{\studio{}} (this work) \\
\midrule
Interface           & Python \textsc{api} / command line          & interactive web application, no code to write \\
Users               & programmers                                 & also non-programmers, across \textsc{nlp}, biology, and the humanities \\
Deployment          & \texttt{pip install}                        & zero install; opens in a browser tab \\
Data locality       & local                                       & local, and never uploaded \\
\addlinespace[3pt]
Execution           & interpreted Python                          & C++ compiled to WebAssembly, on the client \\
Speed ($n{=}2048$)  & $1\times$ (baseline)                        & {$100$ to $2500\times$ faster} \\
vs.\ native C       & --                                          & $6.2\times$ over Biopython (global); within $1.8\times$ of \textsc{ssw} (local) \\
\addlinespace[3pt]
Correctness         & --                                          & bit-exact vs.\ the library and independent refs ($204/204$ per algorithm) \\
\addlinespace[3pt]
Generation metrics  & embedding-based only                        & \textsc{bleu}, chr\textsc{f}, \textsc{rouge}, \textsc{cer}/\textsc{wer} with paired-bootstrap significance \\
Homology search     & named as future work                        & client-side \textsc{blast}, $\le 0.43\%$ off \textsc{ncbi} \textsc{blast}+ \\
Output              & static plots                                & linked interactive figures; export to \textsc{pdf}/\textsc{csv}/\textsc{json}/\textsc{fasta}/\dots \\
\bottomrule
\end{tabular}%
}
\caption{How our Studio extends its own predecessor. The library gathered the string-to-string algorithms behind one Python interface; \studio{} keeps that coverage and adds an interactive, verified, client-side engine that is two to three orders of magnitude faster, together with the generation metrics and homology search the library did not provide, for programmers and non-programmers alike. Speed figures are medians over $30$ random pairs at length $2048$ on an Apple M1 Pro; correctness and \textsc{blast} concordance are defined in~\S~\ref{sec:eval}.}
\label{tab:headtohead}
\end{table*}

\begin{table*}[!h]
\centering
\small
\setlength{\tabcolsep}{5pt}
\renewcommand{\arraystretch}{1.22}
\newcommand{\rr}{\raggedright\arraybackslash}
\begin{tabular}{@{}>{\rr}p{1.85cm}>{\rr}p{5.95cm}>{\rr}p{2.95cm}>{\rr}p{4.1cm}@{}}
\toprule
\textbf{Class} & \textbf{Methods (selected)} & \textbf{Levels} & \textbf{Interactive view} \\
\midrule
Alignment &
\mbox{Needleman--Wunsch}, \mbox{Smith--Waterman}, \mbox{semi-global}, affine (Gotoh), banded, Hirschberg, \textsc{dtw} &
char, word, token, line &
\textsc{dp} matrix + path, dot plot, ribbon, edit script \\
Edit distance &
Levenshtein, Damerau, Hamming, \mbox{Jaro--Winkler}, \textsc{lcs} &
char, word, token &
identity track, redline diff \\
Similarity &
Jaccard, cosine, \textsc{lcs} ratio &
char, word, token &
score readout, method comparison \\
Search &
naive, \textsc{kmp}, \mbox{Boyer--Moore}, \mbox{Rabin--Karp}, \mbox{$k$-mismatch}, \textsc{iupac}, \mbox{both-strand} &
char, word, token &
match map, both-strand track \\
Generation metrics &
\textsc{bleu}, chr\textsc{f}, chr\textsc{f}++, \textsc{rouge}-1/2/\textsc{l}, \textsc{cer}, \textsc{wer}, token-$F_1$ &
word, token, char &
corpus + per-segment, bootstrap \\
Homology &
\texttt{blastn}, \texttt{blastp}, Karlin--Altschul, \mbox{seed-and-extend}, \textsc{msa} + tree &
residue &
hit map, conservation skyline, tree \\
\bottomrule
\end{tabular}
\caption{The string-to-string family as \studio{} implements it: for each operation class, representative
methods, the levels at which they run, and the interactive view the platform provides. Every method
listed is in the released engine and is checked in continuous integration (Section~\ref{sec:eval}).}
\label{tab:taxonomy}
\end{table*}

\end{document}